\documentclass[runningheads]{llncs}
\usepackage{graphicx}

\usepackage{tikz}
\usepackage{comment}
\usepackage{amsmath,amssymb} 
\usepackage{color}
\usepackage{orcidlink}

\usepackage{subcaption}
\usepackage{wrapfig,lipsum,booktabs}
\usepackage[inline]{enumitem}

\usepackage[accsupp]{axessibility}  

\begin{document}
\pagestyle{headings}
\mainmatter
\def\ECCVSubNumber{7127}  

\title{Privacy-Preserving Action Recognition via Motion Difference Quantization} 

\titlerunning{Privacy-Preserving Action Recognition via Motion Difference Quantization}
%
\author{Sudhakar Kumawat\index{Kumawat, Sudhakar} \and
Hajime Nagahara}
\authorrunning{Kumawat and Nagahara}
%
\institute{Osaka University, Japan \\
\email{\{sudhakar,nagahara\}@ids.osaka-u.ac.jp}}

\maketitle

\begin{abstract}
The widespread use of smart computer vision systems in our personal spaces has led to an increased consciousness about the privacy and security risks that these systems pose. On the one hand, we want these systems to assist in our daily lives by understanding our surroundings, but on the other hand, we want them to do so without capturing any sensitive information. Towards this direction, this paper proposes a simple, yet robust privacy-preserving encoder called BDQ  for the task of privacy-preserving human action recognition that is composed of three modules: \textit{\underline{B}lur}, \textit{\underline{D}ifference}, and \textit{\underline{Q}uantization}. First, the input scene is passed to the \textit{Blur} module to smoothen the edges. This is followed by the \textit{Difference} module to apply a pixel-wise intensity subtraction between consecutive frames to highlight motion features and suppress high-level privacy attributes. Finally, the \textit{Quantization} module is applied to the motion difference frames to remove the low-level privacy attributes. The BDQ  parameters are optimized in an end-to-end fashion via adversarial training such that it learns to allow action recognition attributes while inhibiting privacy attributes. Our experiments on three benchmark datasets show that the proposed encoder design can achieve state-of-the-art trade-off when compared with previous works. Furthermore, we show that the trade-off achieved is at par with the DVS sensor-based event cameras. Code available at: \url{https://github.com/suakaw/BDQ_PrivacyAR}

\end{abstract}

\section{Introduction}\label{sec:intro}
For many decades, people have been fascinated with the idea of creating computer vision (CV) systems that can see and interpret the world around them. In today's world, as this dream turns into reality and such systems begin to be deployed in our personal spaces, there is an increased consciousness about ``what'' these systems see and ``how'' they interpret it. Nowadays, we want CV systems that can protect our visual privacy without compromising the user experience. Therefore, there is a growing interest in developing such CV systems that can prevent the camera system from obtaining detailed visual data that may contain any sensitive information, but allow it to capture useful information to successfully perform the CV task \cite{wu2018towards,ryoo2017privacy,tan2020canopic,hinojosa2021learning,raval2017protecting}. For a satisfactory user experience and strong privacy protection, a CV system must satisfy the following properties:
\begin{itemize}
    \item \textbf{Good target task accuracy.} This is necessary for maintaining a good user experience. For example, a privacy-preserving face detection model must detect faces with high precision without revealing facial identity \cite{tan2020canopic}, a privacy-preserving pose estimation model must detect body key-points without revealing  the person identity \cite{hinojosa2021learning}, and an action recognition model must recognize human actions without revealing their identity information \cite{wu2018towards,ryoo2017privacy}. 
    \item \textbf{Strong privacy protection.} Any privacy-preserving model, irrespective of the target task must preserve common visual privacy attributes such as identity, gender, race, color, gait, etc. \cite{orekondy2017towards}. Note that the definition of privacy attributes for a privacy-preserving model may vary depending on its application. Furthermore, strong privacy protection is guaranteed when the model is applied at the point of capture and it is impossible for any adversary to learn or reconstruct the privacy attributes.   
    \item \textbf{Cost-effective and low space-time complexity.} The right to privacy is considered as one of the fundamental rights in the most countries. However, with the increasing availability of low-budget consumer cameras and smartphones, it cannot be fully guaranteed unless the privacy-preserving models are affordable to everyone. This calls for a focus on implementing such models, whether in software or hardware, in a cost-effective manner. Additionally, since the privacy-preserving model needs to be applied at the point of capture in consumer cameras and smartphones with low memory, computation, and power budgets, it must be of low space-time complexity.
\end{itemize}
\begin{figure}[t]
     \centering
     \begin{subfigure}[b]{.15\textwidth}
         \centering
         \includegraphics[width=\textwidth]{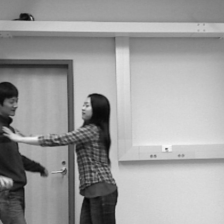}
         \includegraphics[width=\textwidth]{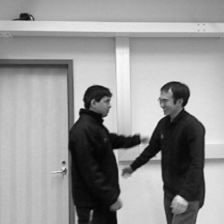}
         \caption*{\scriptsize Orig. Image}
     \end{subfigure}
     \hspace{-.5em}
     \begin{subfigure}[b]{.15\textwidth}
         \centering
         \includegraphics[width=\textwidth]{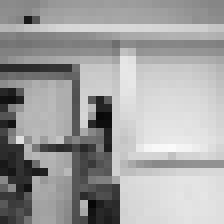}
         \includegraphics[width=\textwidth]{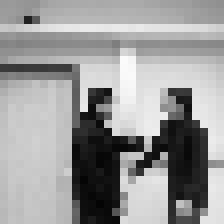}
         \caption*{\scriptsize Downsampled}
     \end{subfigure}
          \hspace{-.5em}
     \begin{subfigure}[b]{.15\textwidth}
         \centering
         \includegraphics[width=\textwidth]{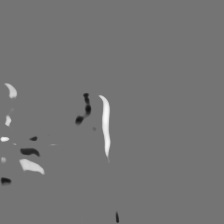}
         \includegraphics[width=\textwidth]{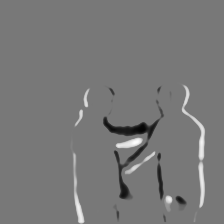}
         \caption*{\scriptsize BDQ}
     \end{subfigure}%
     \hspace{.5em}
     \begin{subfigure}[b]{.15\textwidth}
         \centering
         \includegraphics[width=\textwidth]{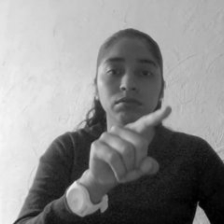}
         \includegraphics[width=\textwidth]{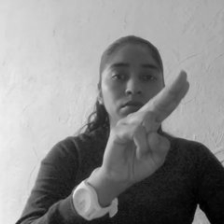}
         \caption*{\scriptsize Orig. Image}
     \end{subfigure}
     \hspace{-.5em}
     \begin{subfigure}[b]{.15\textwidth}
         \centering
         \includegraphics[width=\textwidth]{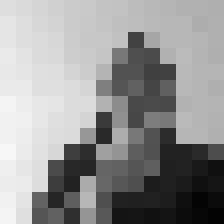}
         \includegraphics[width=\textwidth]{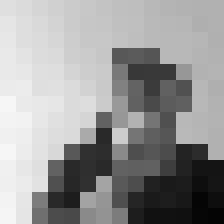}
         \caption*{\scriptsize Downsampled}
     \end{subfigure}
          \hspace{-.5em}
     \begin{subfigure}[b]{.15\textwidth}
         \centering
         \includegraphics[width=\textwidth]{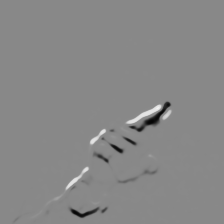}
         \includegraphics[width=\textwidth]{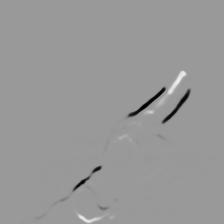}
         \caption*{\scriptsize BDQ}
     \end{subfigure}
        \caption{Left- An example where downsampling is effective for privacy-preserving action recognition. Right- An  example where down-sampling is detrimental for action recognition since the information about the number of fingers is lost. Unlike downsampling, the BDQ preserves both spatial and temporal resolutions.  \label{fig:intro}}
\end{figure}
In this paper, we are interested in developing a privacy-preserving encoder for the task of human action recognition while keeping the above properties in mind. In particular, we develop a simple and robust privacy-preserving encoder called BDQ that is composed of three modules: \textit{Blur}, \textit{Difference}, and \textit{Quantization}. The modules are applied in a sequential manner such that \textit{Blur} module smoothen the edges in the input frames. The \textit{Difference} module computes pixel-wise intensity subtraction between consecutive frames to highlight motion features and suppress high-level privacy attributes. The \textit{Quantization} module removes the low-level privacy attributes from the motion difference frames. The parameters of BDQ are optimized in an end-to-end manner via adversarial training such that it learns to facilitate action recognition while inhibiting privacy attributes. Note that, any successful action recognition model relies on the presence of both spatial and temporal cues in its input for good performance. However, in most cases, applying a privacy-preserving encoder often destroys the spatial resolution which is sometimes detrimental for action recognition as shown in Figure~\ref{fig:intro}.
The design of our BDQ encoder is motivated to alleviate such issues such that it preserves both spatial and temporal cues that are essential for action recognition while discarding attributes that may reveal the privacy information. In summary, the contributions of this work are as follows. 
\begin{itemize}
    \item We propose a simple yet robust privacy-preserving encoder called BDQ for the task of human action recognition. The BDQ encoder allows important spatio-temporal cues for action recognition while preserving privacy attributes at a very low space-time complexity.
    \item We show that the BDQ encoder achieves state-of-the-art trade-off between action recognition and privacy preservation on three benchmark datasets SBU, KTH, and IPN, when compared with other privacy-preserving models. Moreover, we show that the trade-off achieved is at par with the DVS sensor-based event cameras.     
    \item We provide an extensive analysis of the BDQ encoder. The analysis includes an ablation study on the components of BDQ, learning and reconstructing privacy attributes by different adversaries, and a subjective evaluation.  
    \item Finally, we also discuss the feasibility of implementing the BDQ modules using existing hardware (please refer supplementary).
\end{itemize}
\section{Related Work}
In recent years, there is a growing interest in developing privacy-preserving vision systems for various computer vision tasks such as action recognition \cite{wu2018towards,ryoo2017privacy}, face detection \cite{tan2020canopic}, pose estimation \cite{hinojosa2021learning,srivastav2019human}, fall detection \cite{asif2020privacy}, and posture classification \cite{gochoo2020lownet}. Here, we provide a brief overview of  privacy-preserving frameworks for various computer vision tasks, especially human action recognition.

\noindent\textbf{Privacy-preserving computer vision.}
Early privacy-preserving models used hand-crafted features such as blurring, down-sampling, pixelation, and face/object replacement for protecting sensitive information \cite{agrawal2011person,chen2007tools,padilla2015visual}. Unfortunately, such approaches require extensive domain knowledge about the problem setting which may not be feasible in practice. Modern privacy-preserving frameworks take a data-driven approach to hide sensitive information. Such
frameworks learn to preserve privacy via adversarial training \cite{roy2019mitigating} using Deep Neural Networks (DNNs) that train the parameters of an encoder to actively inhibit the sensitive attributes in an visual data against an adversarial DNN whose task is to learn privacy attributes while allowing attributes that are essential for the computer vision task \cite{brkic2017know,pittaluga2019learning,tan2020canopic,wu2018towards,huang2018generative,mirjalili2019flowsan}. Note that the encoder can be a hardware module \cite{hinojosa2021learning,tan2020canopic} or a software module \cite{wu2018towards,roy2019mitigating}. Besides these works, there are other imaging systems that use optical operations to hide sensitive attributes. For example, in  \cite{pittaluga2016pre}, the authors design camera systems that perform blurring and $k$-same face de-identification in hardware. Furthermore, in \cite{wang2019privacy,canh2019deep}, authors explore coded aperture masks to enhance privacy.

\noindent\textbf{Privacy-preserving action recognition.} Early works in this area proposed to learn human actions from low-resolution videos \cite{dai2015towards,ryoo2017privacy,ryoo2018extreme}. Ryoo \textit{et al.} \cite{ryoo2017privacy,ryoo2018extreme}  proposed to learn image transformations to down-sample frames into low-resolution for action recognition. Wang \textit{et al.} \cite{wang2019privacy} proposed a lens-free coded aperture camera system for action recognition that is privacy-preserving. Note that the above frameworks are limited to providing visual privacy and it is not clear, to what extent they provide protection against adversaries like DNNs that may try to learn or reconstruct the privacy attributes. To solve this issue, Ren \textit{et al.} \cite{ren2018learning} proposed to learn a video face anonymizer using adversarial training. The training uses a video anonymizer that modifies the original video to remove sensitive information while trying to maximize the action recognition performance and a discriminator that tries to extract sensitive information from the anonymized videos. Later, Wu \textit{et al.}  \cite{wu2018towards,wu2020privacy} proposed and compared multiple adversarial training frameworks for optimizing the parameters of the encoder function. They used a UNet-like encoder from \cite{johnson2016perceptual} which can be seen as a 2D conv-based frame-level filter. The encoder is trained to allow important spatio-temporal attributes for action recognition, measured by a DNN, and to inhibit the privacy attributes in frames against an ensemble of DNN adversaries whose task is to learn the sensitive information. 
An important drawback of this framework is that the training requires an ensemble of adversaries in order to provide strong privacy protection. Concurrent to our work, Dave \textit{et al.} \cite{dave2022spact} proposed a self-supervised framework for training a UNet-based privacy-preserving encoder.

\section{Proposed Framework}\label{sec:bdq}
In this section, we present the design of our BDQ encoder which is composed of a series of modules. We also discuss an adversarial training scheme to optimally train the parameters of these modules for the two seemingly contradictory tasks: protecting privacy and enabling action recognition. 

\subsection{BDQ: \underline{B}lur \underline{D}ifference \underline{Q}uantization}
\begin{figure}[t]
\centering
\includegraphics[width=.9\textwidth]{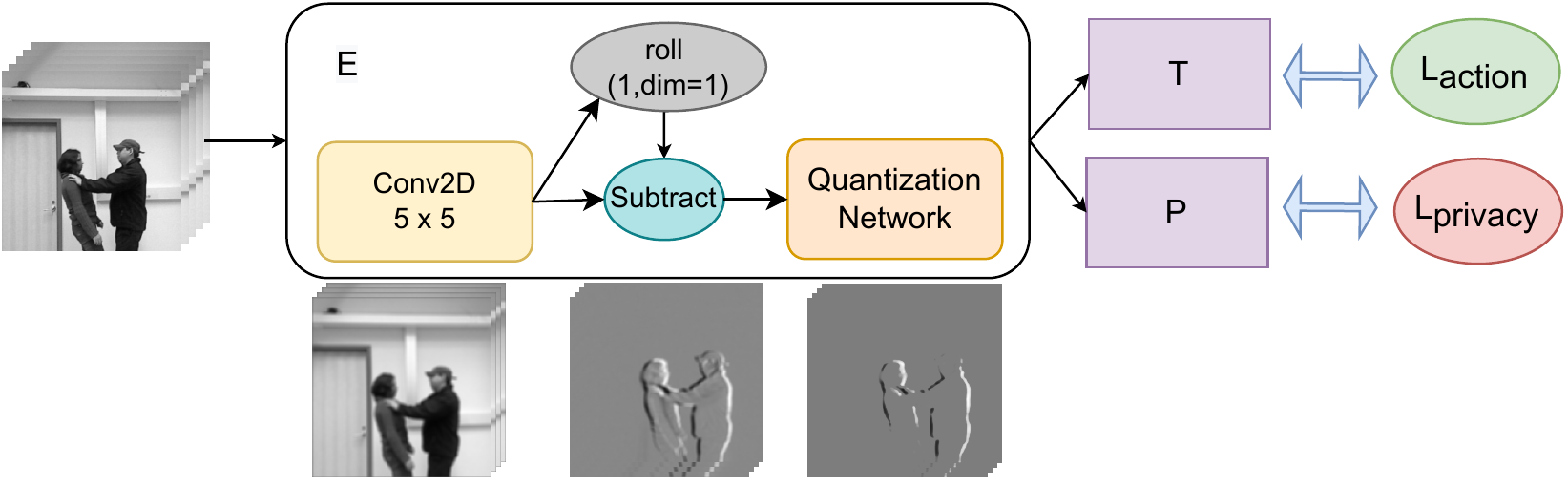}
\caption{ The BDQ encoder architecture and adversarial training framework for privacy-preserving action recognition. $roll(1,dim=1)$ operation shifts frames by one along the temporal dimension.  \label{fig:bdq}}
\end{figure}

The BDQ encoder, as the name suggests is composed of three modules: \textit{Blur}, \textit{Difference}, and \textit{Quantization}. Given a scene, the three modules are applied to it in a sequential manner as shown in Figure~\ref{fig:bdq}. Here, for each module, we provide a detailed description of its implementation and role in preserving privacy and enabling action recognition.  

\textbf{Blur.} The goal of the \textit{Blur} module is to blur the spatial edges while preserving important spatial features for action recognition. More importantly, its task is to suppress the obvious privacy features that may leak at the spatial edges in motion difference frames that will be produced by the \textit{Difference} module. Given a frame $v_i$, we define a video as $V=\{v_i|i=1,2,..,t\}$ where $t$ is the number of frames. The blurred frame $B_{v_{i}}$ is convolution of video frame $v_i$ and a 2D Gaussian kernel $G_\sigma$ of $\sigma$ standard deviation and defined as
$B_{v_{i}} = G_\sigma v_i$,   where  $G_\sigma = \frac{1}{2\pi\sigma^2}exp(-\frac{x^2+y^2}{2\sigma^2})$. Window-size of the kernel is kept as $5\times 5$ and $\sigma$ is learned during the adversarial training. A small window-size is chosen since it stabilizes training and avoids losing important spatial features. 

\textbf{Difference.} Given two consecutive frames from an action video after passing through the \textit{Blur} module, this module performs pixel-wise numerical subtraction between their intensity values and outputs a single frame as $D(B_{v_{i}},B_{v_{j}}) = B_{v_{i}}-B_{v_{j}}$.
It serves two important purposes: First, it highlights the motion features between the two frames by bringing out the direction of motion which improves action recognition accuracy. This is evident from the fact that many current state-of-the-art action recognition methods such as  \cite{jiang2019stm,li2020tea,liu2020teinet,wang2021tdn} use such temporal-difference modules in the feature space to develop action recognition models. Second, it suppresses obvious high-level spatial privacy cues which helps in preserving privacy. Note that this module does not contain any learnable parameters. Furthermore, when implemented for online applications where frames are continuously being captured, it needs to store a copy of the previous frame for pixel-wise numerical subtraction.     

\textbf{Quantization.} Although, the \textit{Blur} and the \textit{Difference} modules contribute to suppressing high-level spatial privacy cues, they cannot fully protect against adversaries (see Section~\ref{sec:analysis_ablation}). This is because they still allow low-level spatial privacy cues that are good enough for learning and reconstructing the privacy attributes (see Section~\ref{sec:analysis_reconst}). To alleviate this issue, the task of the \textit{Quantization} module is to remove such information by applying a pixel-wise quantization function on the motion difference frames that are output by the \textit{Difference module}. Conventionally, a quantization function is defined by $y=\sum_{n=1}^{N-1} \mathcal{U}(x-b_{i})$, where $y$ is the discrete output, $x$ is a continuous input, $b_i = \{0.5, 1.5, 2.5,\ldots, N - 1.5\}$, $N=2^{k}$, $k$ is the number of bits, and $\mathcal{U}$ is the Heaviside function. Unfortunately, such a formulation is not differentiable and therefore not suitable for back-propagation. Thus, following \cite{yang2019quantization,tan2020canopic}, we approximate the above quantization function with a differentiable version by replacing the Heaviside function with the sigmoid $\sigma()$ function, which results in the following formulation: $\sum_{n=1}^{N-1} \sigma(H(x-b_{i}))$, where $H$ is a scalar hardness term. Here, the parameters learned are the $b_{i}$ values. In this work, we fix the number of $b_{i}$ values to be 15 and initialize them to the values $0.5, 1.5,\ldots, 14.5$. Furthermore, the input to the \textit{Quantization} module is normalized to have values between 0 to 15. Note that, here we allow quantization output to be non-integer since no hardware constrain is imposed on BDQ. 

\subsection{Training BDQ Encoder}\label{sec:bdq_train}
The BDQ encoder contains several parameters such as the standard deviation of the \textit{Blur} module and the intervals of the \textit{Quantization} module. Our goal is to set these parameters such that: 
\begin{enumerate*}[label=(\arabic*)]
   \item the privacy attributes cannot be learned from the BDQ output by any adversary,
   \item recognizing action attributes must be feasible with high precision. 
\end{enumerate*}
Although, we can achieve this goal by setting these parameters heuristically, however, it has been shown that a better performance can be achieved if they are learned in a data-driven fashion 
\cite{tan2020canopic,wu2018towards,hinojosa2021learning}.

Our training framework consists of three components: \begin{enumerate*}[label=(\arabic*)]
   \item the BDQ encoder denoted as $E$,
   \item a 3D CNN for predicting target action attribute denoted as $T$,
   \item and a 2D CNN for predicting privacy attribute denoted as $P$.
\end{enumerate*}
The three components are connected as illustrated in Figure~\ref{fig:bdq} such that the output of $E$ is simultaneously passed to the networks  $T$  and  $P$.   Following \cite{roy2019mitigating}, the parameter optimization of $E$, $T$, and $P$ is formulated as a three-player non-zero sum game where the goal of $E$ is to maximize the likelihood of the target action attributes,  measured by $E$, while maximizing the uncertainty in the privacy attributes, measured by $P$. Such a training procedure consists of two steps that are iterated until the privacy attributes are sufficiently preserved without a significant compromise in action recognition. In the first step, $P$ is fixed and $E$ and $T$  are trained together using the following loss function.
\begin{equation}
  \mathcal{L}(V,\theta_E,\theta_T)=\mathcal{XE}(T(E(V)), L_{action})- \alpha\mathcal{E}(P(E(V))) 
\end{equation}
Here, $\mathcal{XE}$ and $\mathcal{E}$ refer to the cross-entropy loss and entropy function, respectively. $\theta_E$ and $\theta_T$ denote parameters of $E$ and $T$, respectively. $L_{action}$ is the ground-truth action label and $\alpha$ is the adversarial weight that allows a trade-off between action and privacy recognition. In the second step, $E$ and $T$ are fixed, and $P$ is trained using the following loss function. Here, $\theta_P$ denote the parameters of $P$ and $L_{privacy}$ are the ground-truth privacy labels.
\begin{equation}
  \mathcal{L}(V,\theta_P)=\mathcal{XE}(P(E(V)), L_{privacy})
\end{equation}

\section{Experiments}\label{sec:experiments}
\subsection{Datasets}
\textbf{SBU.} The SBU Kinect Interaction Dataset \cite{yun2012two} is a two-person interaction dataset for video-based action recognition, recorded at 15 fps. It consists of seven actors interacting in pairs in the following eight ways: approaching, departing, pushing, kicking, punching, exchanging objects, hugging, and shaking hands. Originally, the dataset comes divided into 21 sets such that each set corresponds to a pair of actors performing all the eight interactions. Furthermore, the 21 sets are created such that the same two actors may appear in two different sets. In such a case, in the first set, one actor acts and the other reacts and vice versa in the second set. For example, in set s01s02, actor 1 is acting and actor 2 is reacting; similarly, in set s02s01, actor 2 is acting and actor 1 is reacting. Since, such sets contain the same pair of actors, they can be combined into one class. Following this procedure, we reduce the number of 21 original sets to 13 different distinct actor-pair sets. For our setting, given a video, the target task is to classify it into one of the eight interaction/action classes while the privacy label prediction task is to recognize the actor-pair among the 13 actor-pairs. Note the above method is identical to the one followed in \cite{wu2018towards,wu2020privacy} for developing their privacy-preserving action recognition framework.

\noindent\textbf{KTH.} The KTH dataset~\cite{schuldt2004recognizing} is a video-based action recognition dataset, recorded  at 25 fps. It consists of 25 actors, each performing the following six actions: walk, jog, run, box, hand-wave, and hand clap. The different actions are recorded in different settings and variations including outdoor, outdoor with scale variation, outdoor with different clothes, and indoor. In our experiments, we use the six action classes for the action recognition task and the 25 actor identities for the privacy label prediction task. 

\noindent\textbf{IPN.} The IPN hand gesture dataset~\cite{benitez2021ipn} is a video-based hand gesture dataset, recorded at 30 fps. It consists of 50 actors, each performing the following 13 hand gestures that are common in interacting
with a touch-less screen: pointing with one finger, pointing with two fingers, click with one finger, click with two fingers, throw up, throw down, throw left, throw right, open twice, double click with one finger, double click with two fingers, zoom in, and zoom out. In our experiments, we use the 13 hand gesture classes for the action recognition task and the gender (male/female, 2 classes) of the actors for the privacy label prediction task.

\subsection{Implementation}\label{sec:implementation}
\noindent\textbf{Adversarial Training.}  Our adversarial training framework consists of three components: \begin{enumerate*}[label=(\arabic*)]
   \item the BDQ encoder,
   \item an action recognition model which is set to a 3D ResNet-50 network,
   \item and a privacy attribute prediction model which is set to a 2D ResNet-50 network.
\end{enumerate*} Furthermore, the 3D ResNet-50 and the 2D ResNet-50 networks are initialized with Kinetics-400 and ImageNet pre-trained weights, respectively.
For training, we densely sample $t$ consecutive  frames ($t=16$ for SBU, $t= 32$ for KTH and IPN) from the input video to form an input sequence. For spatial data augmentation, we randomly choose for each input sequence, a spatial position and a scale to perform a multi-scale cropping where the scale is picked from the set $\{1,\frac{1}{2^{1/4}},\frac{1}{2^{3/4}},\frac{1}{2}\}$. The final output is an input sequence with size $224\times 224$. As shown in Figure~\ref{fig:bdq}, the input sequence is then passed through the BDQ encoder whose output is then passed to the 3D ResNet-50 model for action recognition and  2D ResNet-50 model for predicting privacy labels. The optimization of parameters of the three networks is done according to the adversarial training framework discussed in Section~\ref{sec:bdq_train} with $\alpha=2,1,$ and $8$, for SBU, KTH, and IPN, respectively. Scaler hardness $H$ is set to $5$ for all datasets. The adversarial training is performed for 50 epochs with SGD optimizer,
$lr=0.001$, cosine annealing scheduler, and batch size of 16.

\noindent\textbf{Validation.}  We freeze the trained BDQ encoder and  newly instantiate a 3D ResNet-50 model for action recognition and a 2D ResNet-50 model for privacy label prediction. Note that the 3D ResNet-50 and the 2D ResNet-50 models are initialized with Kinetics-400 and ImageNet pre-trained weights, respectively. We use the BDQ encoder output on the train set videos to train the 3D ResNet-50 model for action recognition and the 2D ResNet-50 model for predicting privacy labels. Both the networks are trained for 50 epochs with SGD optimizer, $lr=0.001$, cosine annealing scheduler, and batch size 16. For validation, we sample consecutive $t$ frames ($t=16$ for SBU, $t= 32$ for KTH and IPN) from each input video without any random shift, producing an input sequence. We then center crop (without scaling) each frame in the sequence with a square region of size $224\times 224$. For action recognition, we use the generated sequence on the 3D ResNet model to report the clip-1 crop-1 accuracy. For privacy prediction, we  average the softmax outputs by the 2D ResNet-50 model over $t$ frames and report the average accuracy.

\begin{figure}[t]
    \centering
     \begin{subfigure}[b]{.32\textwidth}
     \centering
     \includegraphics[width=\textwidth]{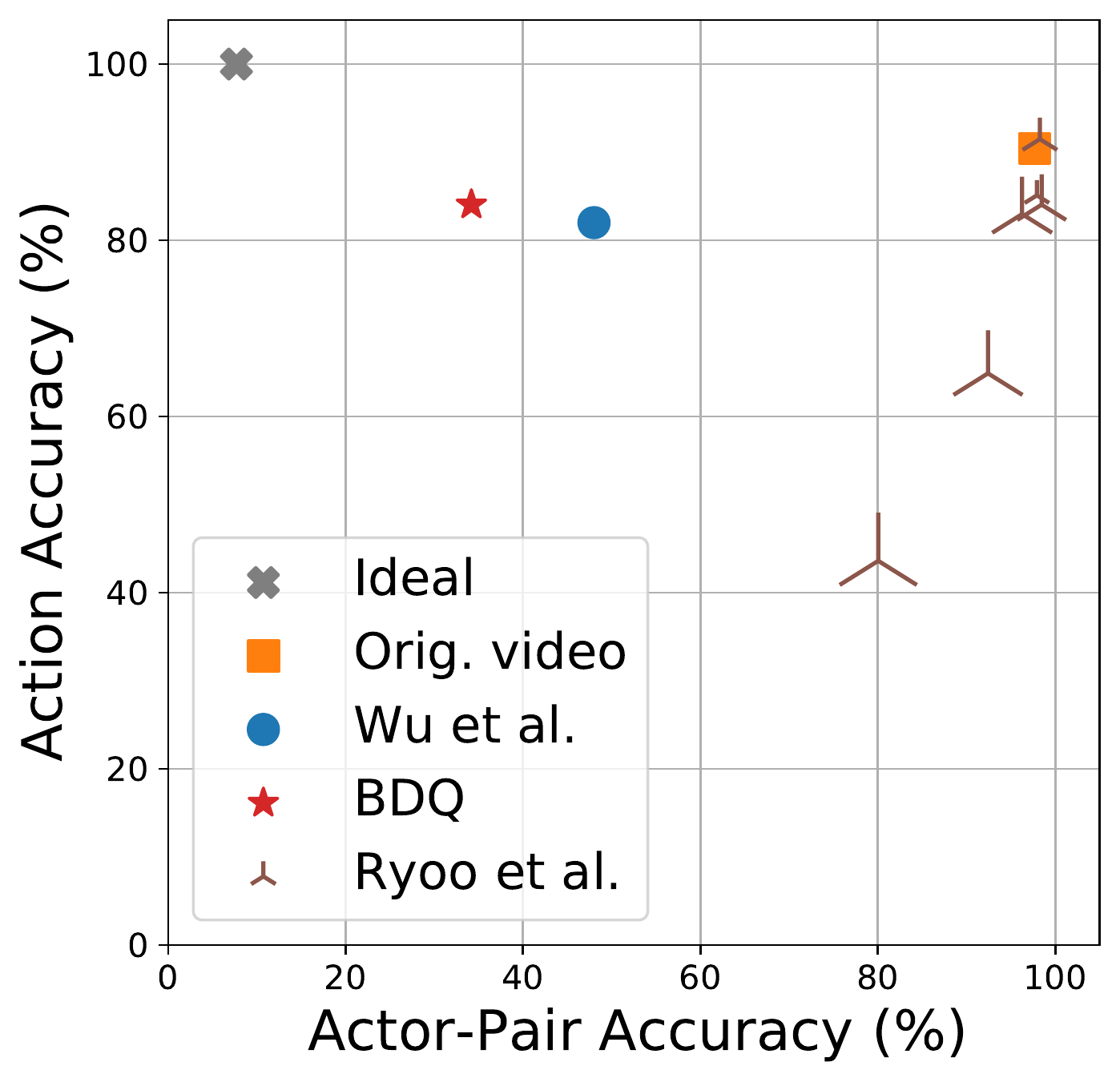}
     \includegraphics[width=\textwidth]{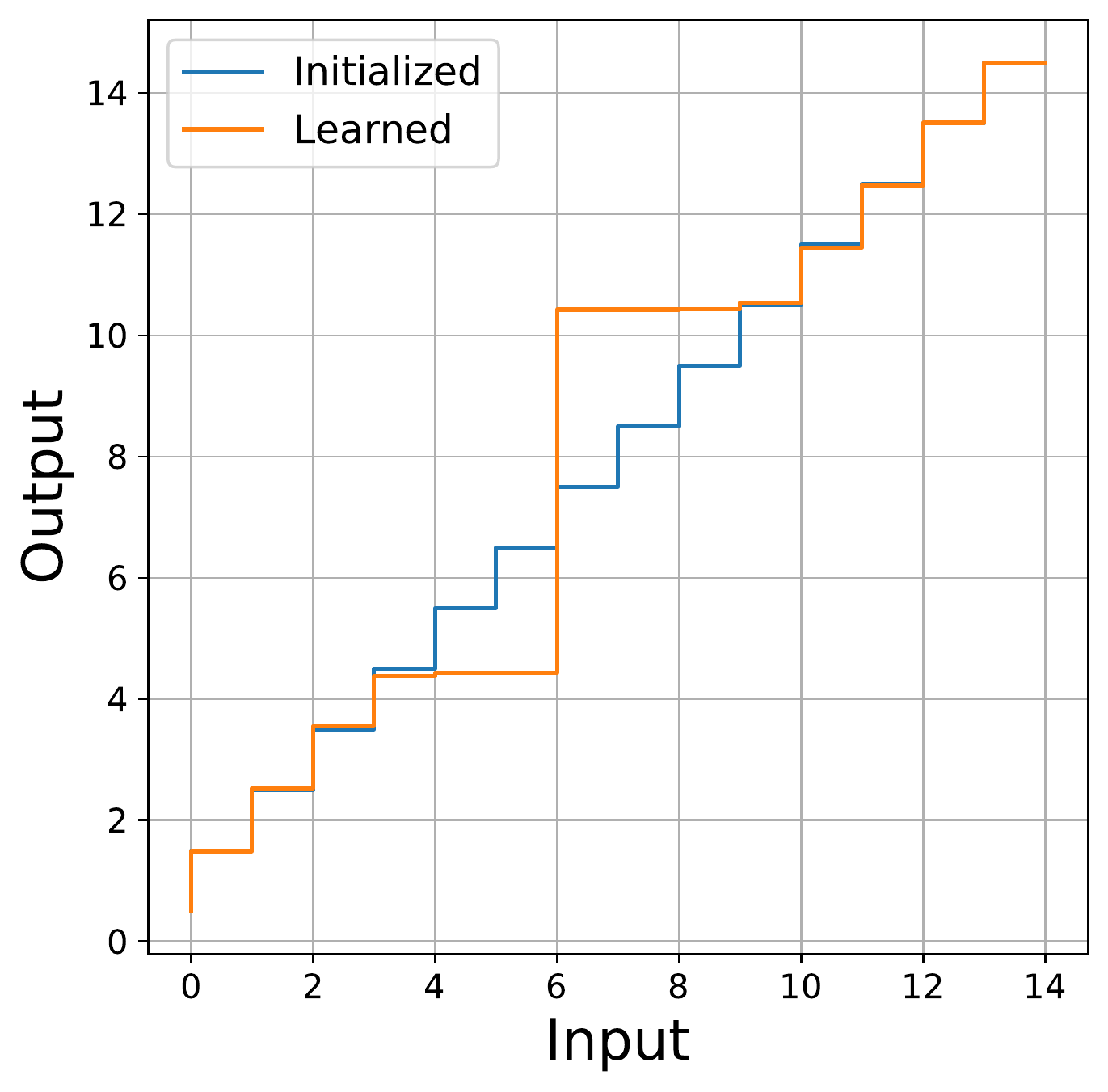}
     \caption*{\scriptsize SBU}
     \end{subfigure}
     \hspace{.1em}
     \begin{subfigure}[b]{.32\textwidth}
     \centering
     \includegraphics[width=\textwidth]{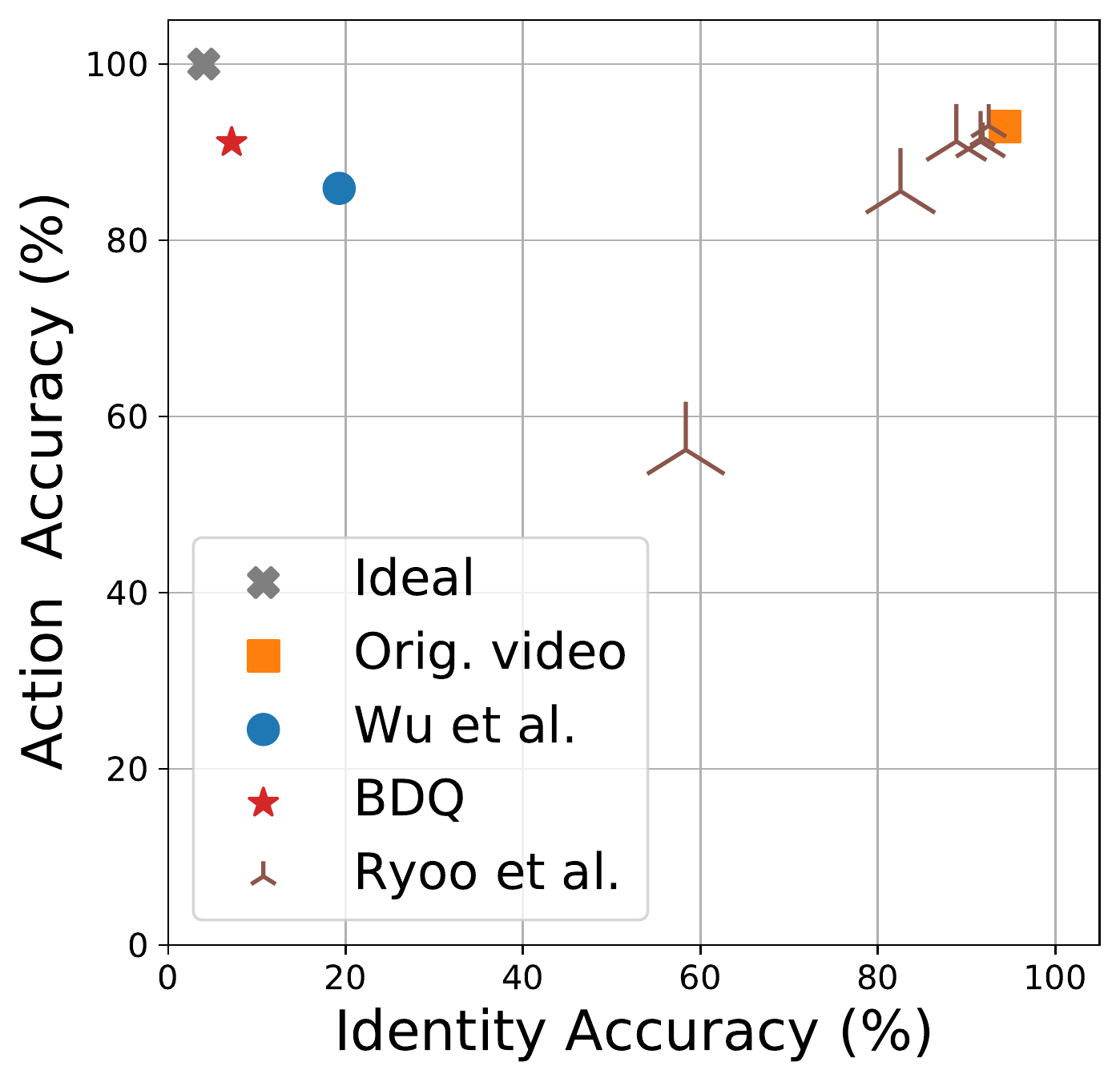}
     \includegraphics[width=\textwidth]{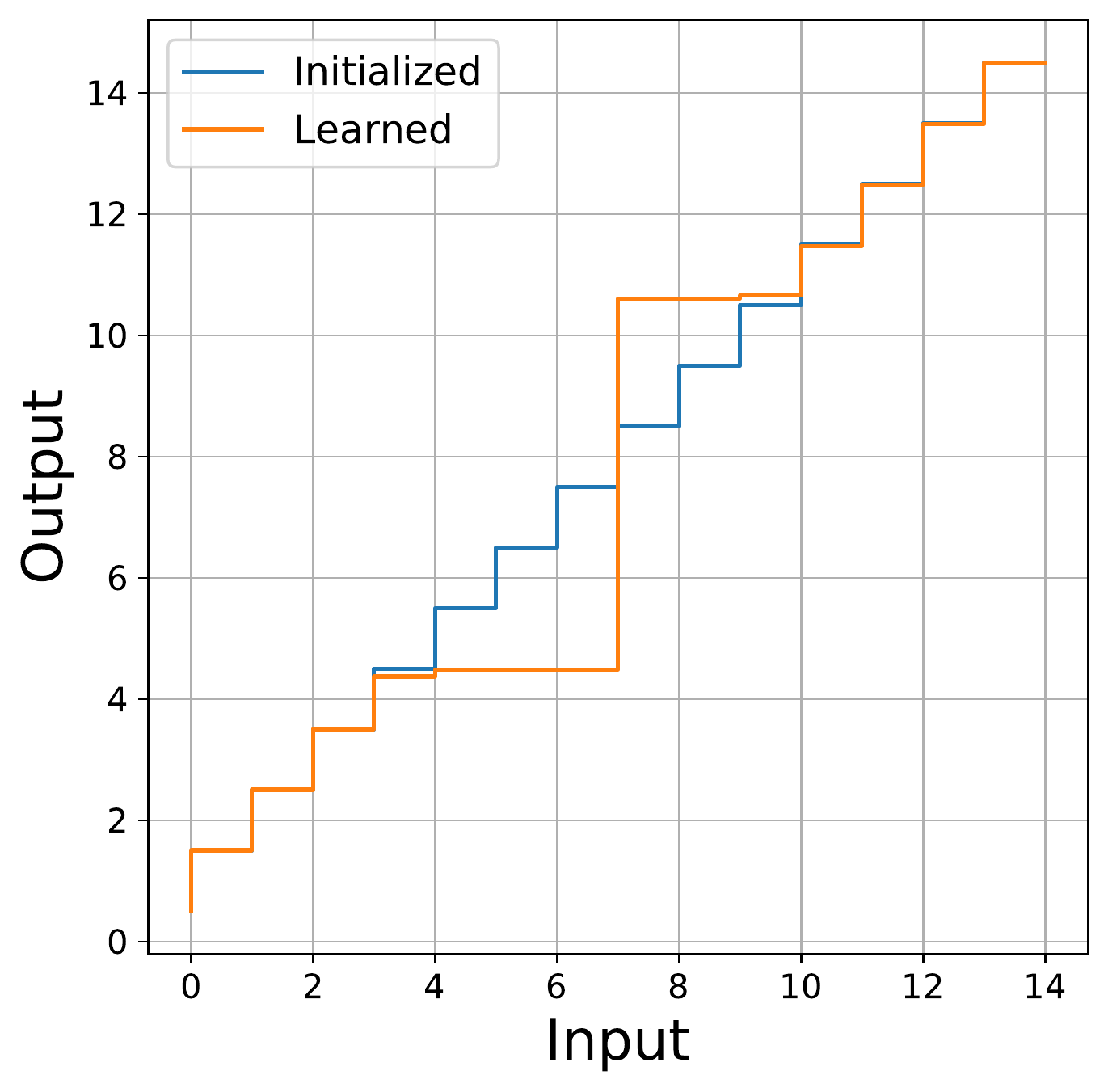}
     \caption*{\scriptsize KTH}
     \end{subfigure}
     \hspace{.1em}
     \begin{subfigure}[b]{.32\textwidth}
     \centering
     \includegraphics[width=\textwidth]{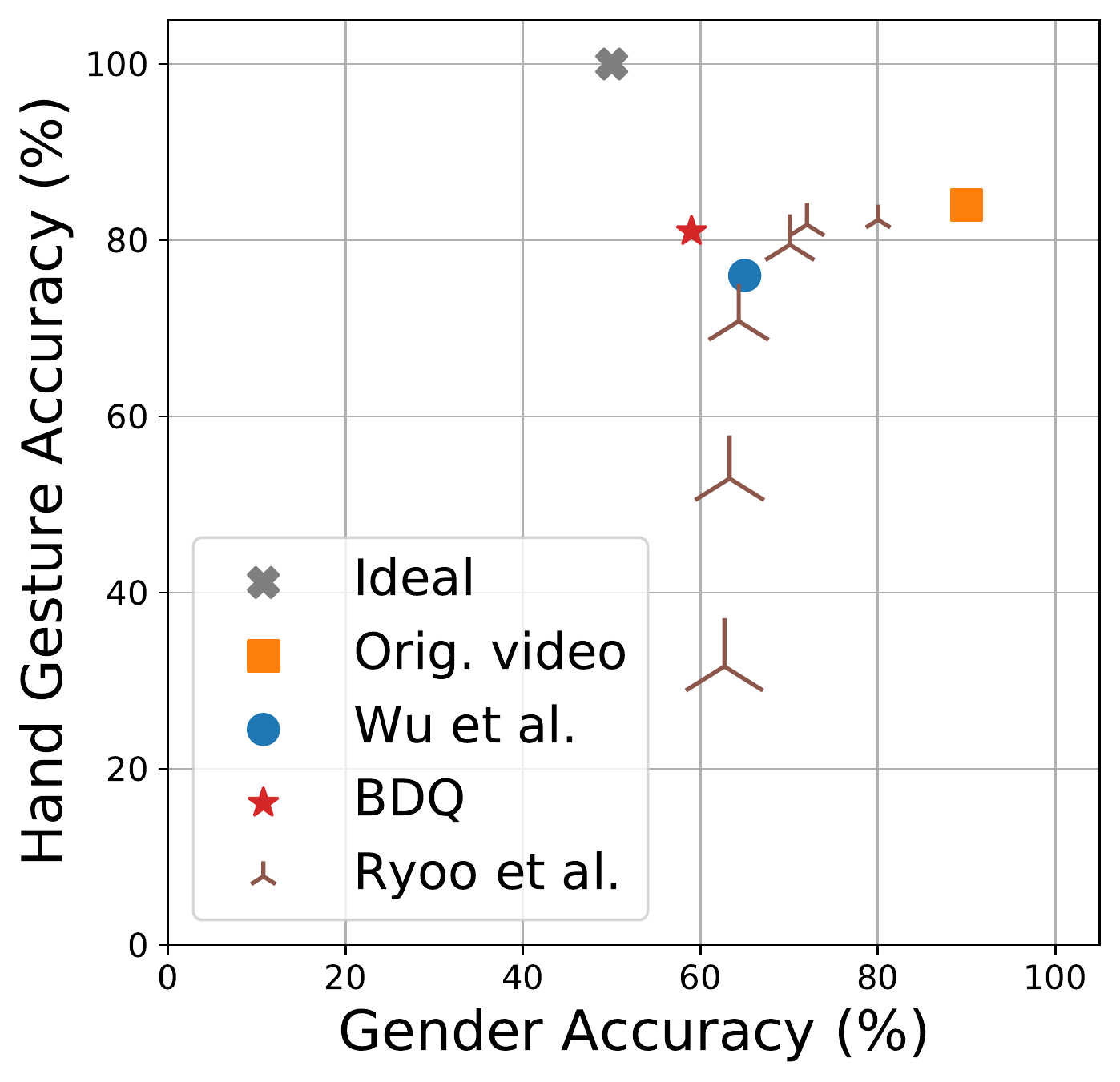}
     \includegraphics[width=\textwidth]{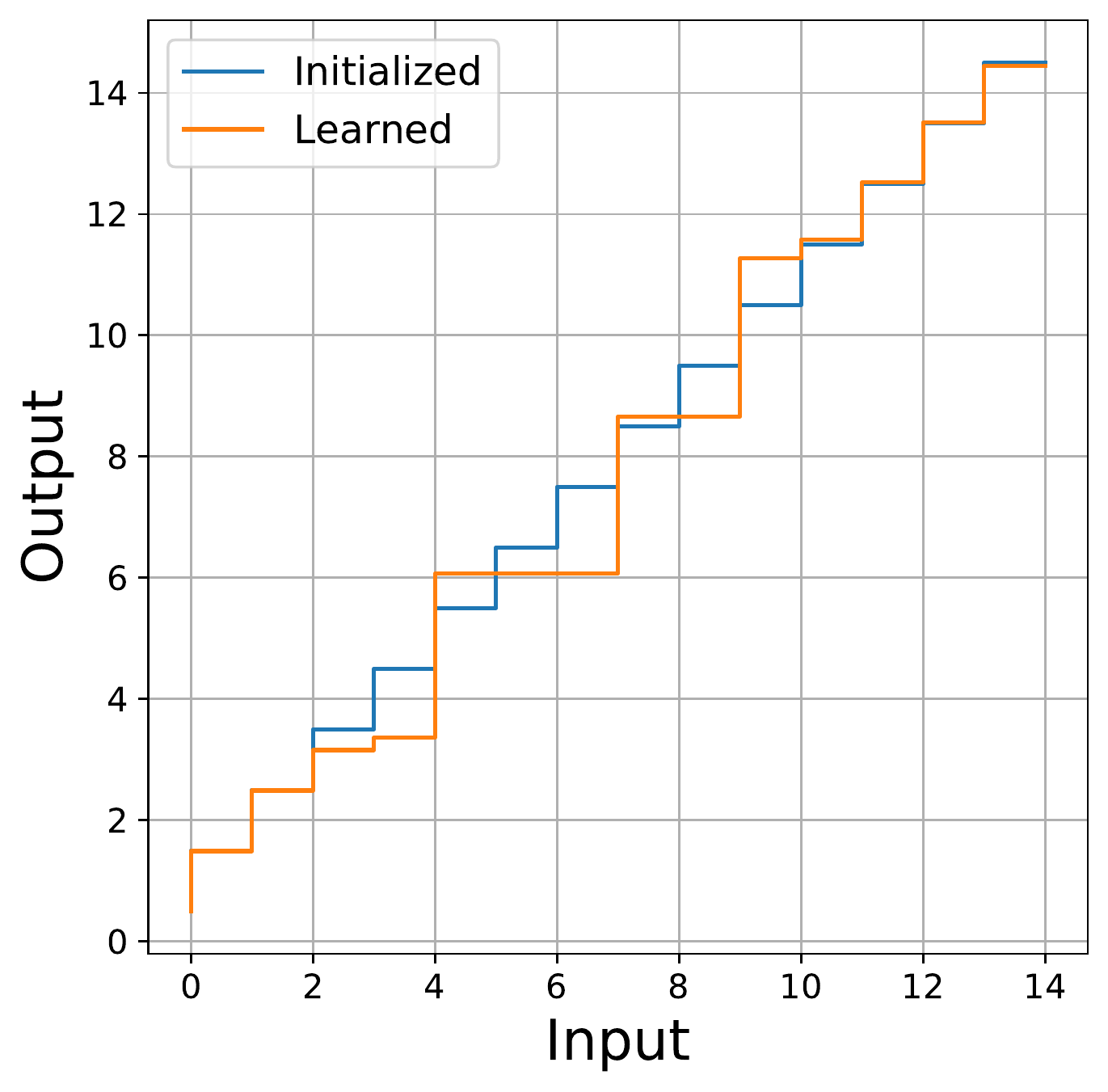}
     \caption*{\scriptsize IPN}
     \end{subfigure}
 \caption{Performance trade-off (row~1) and learned quantization steps (row~2) on the three datasets: SBU, KTH, and IPN.  \label{fig:results_all}}
\end{figure}

\subsection{Results}\label{sec:results}
Figure~\ref{fig:results_all} presents our results on the three datasets. We use the visualization proposed in \cite{wu2018towards} to illustrate the trade-off between the action/gesture recognition accuracy and the privacy label prediction accuracy. We compare our method with two different methods for preserving  privacy in action videos: Ryoo \textit{et al.}~\cite{ryoo2017privacy} and Wu \textit{et al.}~\cite{wu2020privacy}. 

In Ryoo \textit{et al.}~\cite{ryoo2017privacy}, the degradation encoder is a down-sampling module. Here, a high-resolution action video is down-sampled into multiple videos of a fixed low resolution by applying different image transformations that are optimized for the action recognition task. These transformations include sub-pixel translation, scaling, rotation, and other affine transformations emulating possible camera motion. The low-resolution videos can then be used for training action recognition models. In our experiments, we consider the following low spatial resolutions: $112\times 112$, $56\times 56$, $28\times 28$, $14\times 14$, $7\times 7$, and $4\times 4$. For each resolution, we generate corresponding low-resolution videos by applying the learned transformations and train a 3D ResNet-50 model for action recognition and a 2D ResNet-50 model for predicting privacy labels. Figure~\ref{fig:results_all} row~1 reports the results of this experiment where a bigger marker depicts a larger down-sampling rate. We observe that, as the down-sampling rate increases, the privacy label prediction accuracy drops for all the datasets. Unfortunately, the action recognition accuracy drops at a faster rate which is contradictory to the desired behavior.

\begin{wraptable}{R}{0.37\textwidth}
	\scriptsize
	\begin{tabular}[width=.9\textwidth]{lccc}
		\toprule
		\textbf{Method} & \textbf{Params.}	& \textbf{Size}  & \textbf{FLOPs} \\
		\midrule
		Wu \textit{et al}  &  1.3M	& 3.8Mb &  166.4G 	\\
		BDQ		&  16	& 3.4Kb	& 120.4M	\\
		\bottomrule
	\end{tabular}
	\caption{Comparison of space-time complexity of BDQ and Wu \textit{et al}.}
	\label{tab:spacetime}
\end{wraptable}

In Wu \textit{et al}~\cite{wu2020privacy}, the degradation encoder is a UNet-like network that is implemented using the model from \cite{johnson2016perceptual}. It can be seen as a 2D convolution-based frame-level filter and converts each frame into a feature map of the same shape as the original frame. For training the encoder, we use the same adversarial training method as described in Section~\ref{sec:bdq_train} which is same as the original work. However, the privacy model is reset after certain iterations to improve performance as done in the original work. For a fair comparison, the adversarial training is performed using a 3D ResNet-50 model for action recognition and a 2D ResNet-50 model for predicting privacy labels. Furthermore, the validation method followed is identical to the one mentioned in Section~\ref{sec:implementation}. Figure~\ref{fig:results_all} row~1 reports our evaluation results of these experiments. We observe that, in comparison to Ryoo \textit{et al.}~\cite{ryoo2017privacy},  Wu \textit{et al.}~\cite{wu2020privacy} performs significantly better across all the datasets in terms of preserving privacy and enabling action recognition. 

Our proposed BDQ encoder surpasses both Ryoo \textit{et al.}~\cite{ryoo2017privacy} and Wu \textit{et al.}~\cite{wu2020privacy} by a significant margin across all the datasets in preserving privacy and enabling action recognition. As seen in Figure~\ref{fig:results_all} row~1, it is closer to the ideal trade-off than any other method. Finally, Table~\ref{tab:spacetime} compares the space-time complexity of the BDQ and Wu \textit{et al} encoders.  We observe that BDQ uses significantly less parameters and computation in comparison to the Wu \textit{et al} encoder.

\section{Analysis}
\subsection{Ablation study}\label{sec:analysis_ablation}
\begin{wrapfigure}{R}{0.66\textwidth}
\centering
\includegraphics[width=0.33\textwidth]{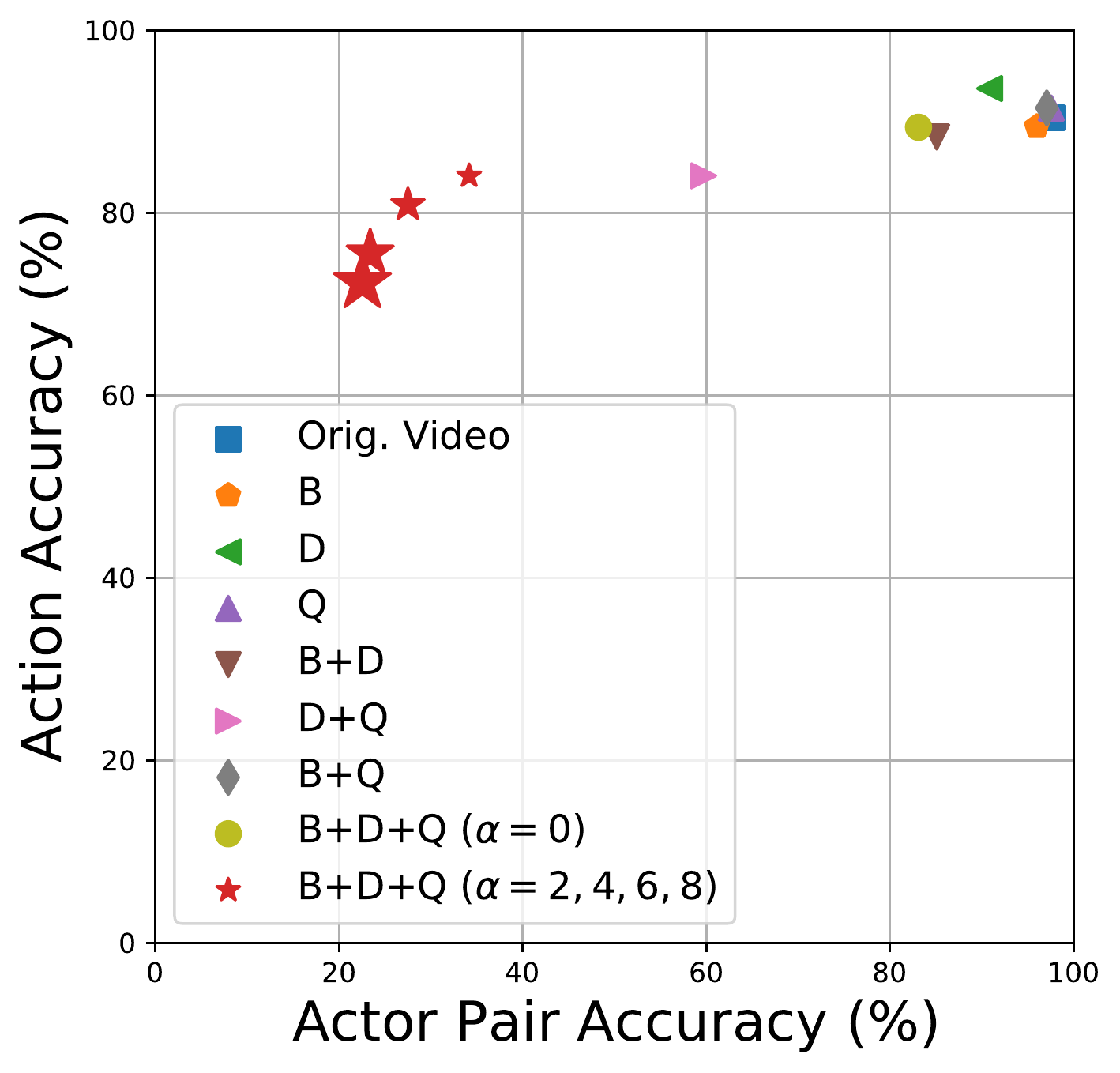}%
\includegraphics[width=0.315\textwidth]{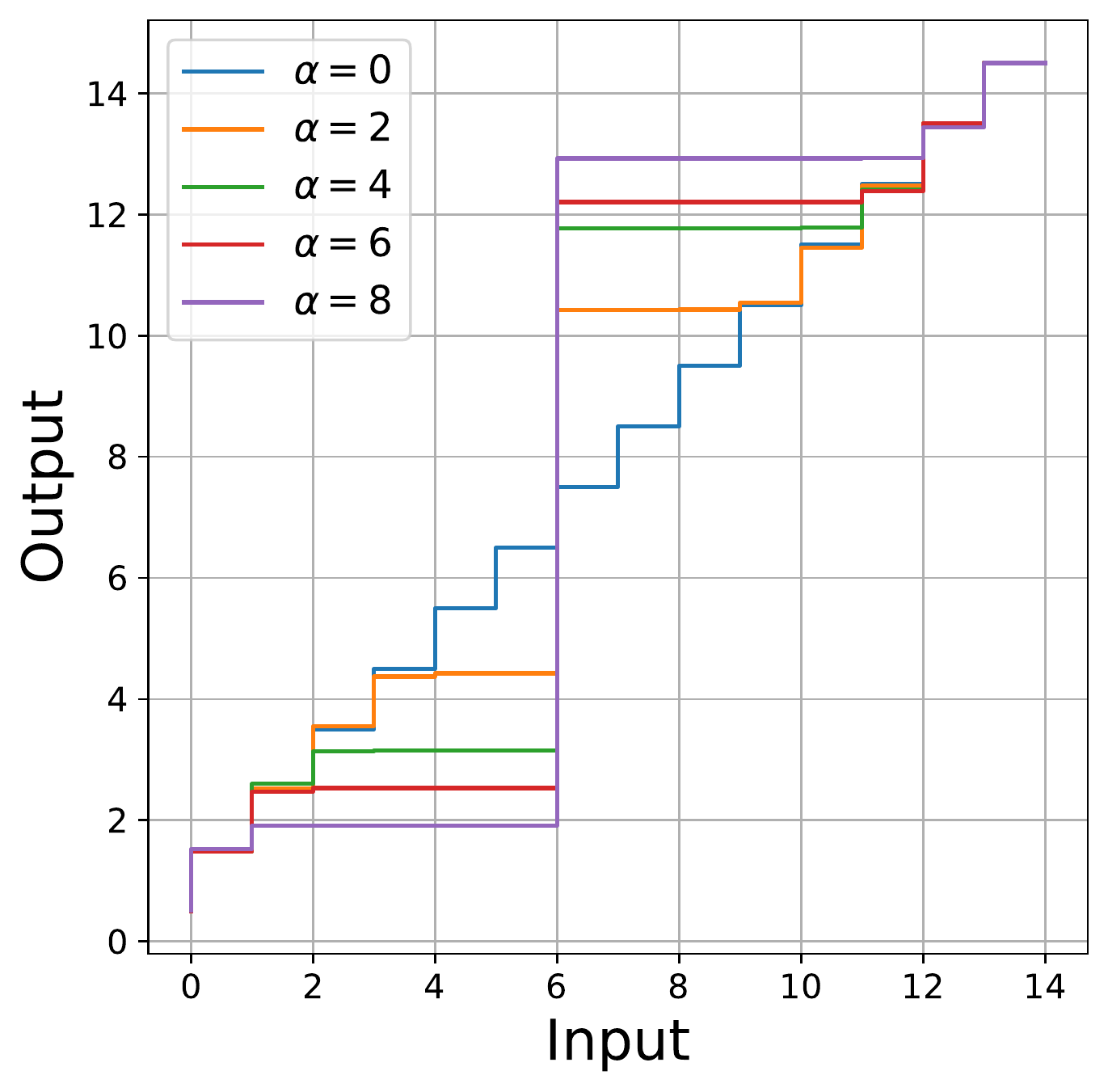}
\caption{\label{fig:frog1} Left- Results of the ablation study. Here, a bigger $\textcolor{red}{\bigstar}$ corresponds to a higher value of $\alpha$. Right- Effect of the adversarial parameter $\alpha$ on the quantization steps. \label{fig:ablation}}
\end{wrapfigure}
As described in Section~\ref{sec:bdq}, the BDQ encoder consists of three modules: \textit{Blur} (B), \textit{Difference} (D), and \textit{Quantization} (Q). Here, we study the role of each of these modules in preserving privacy and enabling action recognition. For this, we take the pre-trained BDQ encoder that was learned on the SBU dataset in Section~\ref{sec:experiments} and select various combinations of its modules to study their contribution to action recognition and actor-pair recognition. For each combination, we freeze the parameters of its module(s) and use its output to train a 3D ResNet-50 (pre-trained on Kinetics-400) for action recognition and a 2D ResNet-50 (pre-trained on ImageNet) for actor-pair recognition. Note that, for both the networks, the training, and the validation procedures are identical to the one used  in Section~\ref{sec:experiments}. 
Figure~\ref{fig:ablation} provides results of this study on the SBU dataset. We observe that the combinations `B', `D', `Q', `B+D', and `B+Q', have very little effect in preserving privacy information and achieve results close to the case when original video is used. Interestingly, the combination `D' achieves a higher action recognition accuracy among all the combinations, signifying its ability to produce better temporal features. Furthermore, a good drop in privacy accuracy is observed when `D' and `Q' are used together. Moreover, this accuracy further drops drastically when all `B', `D', and `Q' are used together. Finally, we also study the effect of the adversarial parameter $\alpha$ on these modules as shown in Figure~\ref{fig:ablation}. We observe that when no adversarial training is performed, i.e $\alpha=0$, there is a very little drop in privacy accuracy. However, with the increase in the value of $\alpha$, both action recognition and actor-pair accuracy begin to fall, with the action recognition accuracy falling more sharply. Figure~\ref{fig:ablation} presents the learned quantization values corresponding to each value of $\alpha$. We observe that with the increase in $\alpha$ the amount of quantization increases which leads to drop in action and actor-pair accuracies. Please refer to supplementary for more studies.

\subsection{Strong Privacy Protection}\label{sec:analysis_privacy}
\begin{figure}[t]
\centering
\includegraphics[width=.7\textwidth]{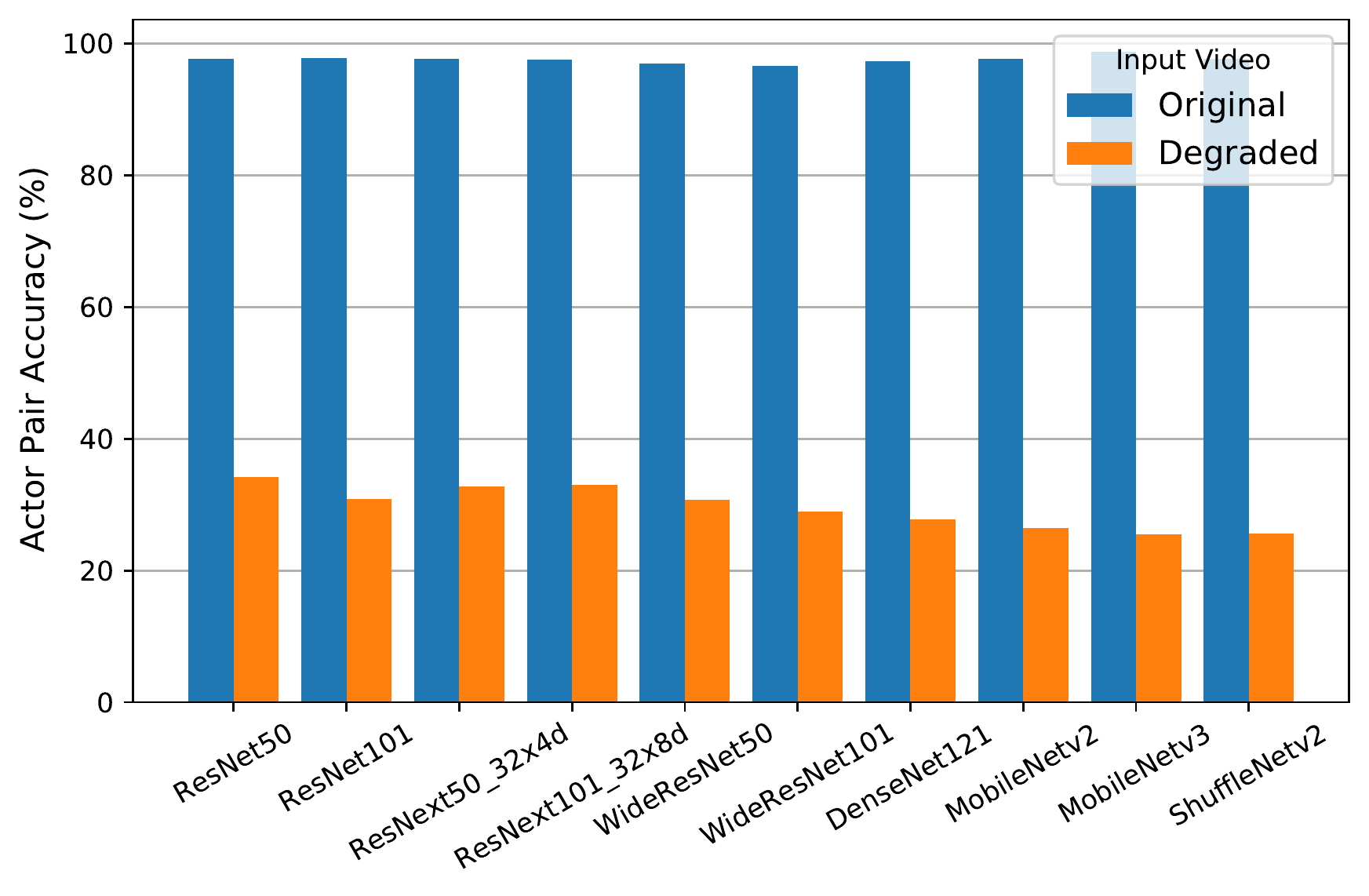}
\caption{Actor-pair accuracy on various image classification networks.\label{fig:analysis_privacy}}
\end{figure}

A significant challenge for any privacy-preserving model is to provide protection against any possible, seen and unseen adversary, that may try to learn the privacy information. In order to show that our proposed framework provides such strong privacy protection, we prepare a list of ten state-of-the-art image classification networks (adversaries) as shown in Figure~\ref{fig:analysis_privacy}. We take the pre-trained BDQ encoder from Section~\ref{sec:experiments} and use its output (degraded video) to train the above networks for predicting the actor-pair labels on the SBU dataset. Furthermore, we also train these networks on the original videos to prepare corresponding baselines for comparison.  Note that all the networks are initialized with ImageNet pre-trained weights, and the training and inference procedures are identical to the one used for the adversary in Section~\ref{sec:experiments}. From Figure~\ref{fig:analysis_privacy}, we observe that the BDQ encoder consistently protects privacy information against all the networks with ResNet-50~\cite{he2016deep} performing the best at 34.18\% and MobileNet-v3~\cite{howard2019searching} performing the worst at 25.46\%. 
Note that, among all the adversaries, the BDQ encoder had only seen ResNet-50 during its training.  
\subsection{Generalized Spatio-temporal Features}
\begin{figure}[t]
\centering
\includegraphics[width=.7\textwidth]{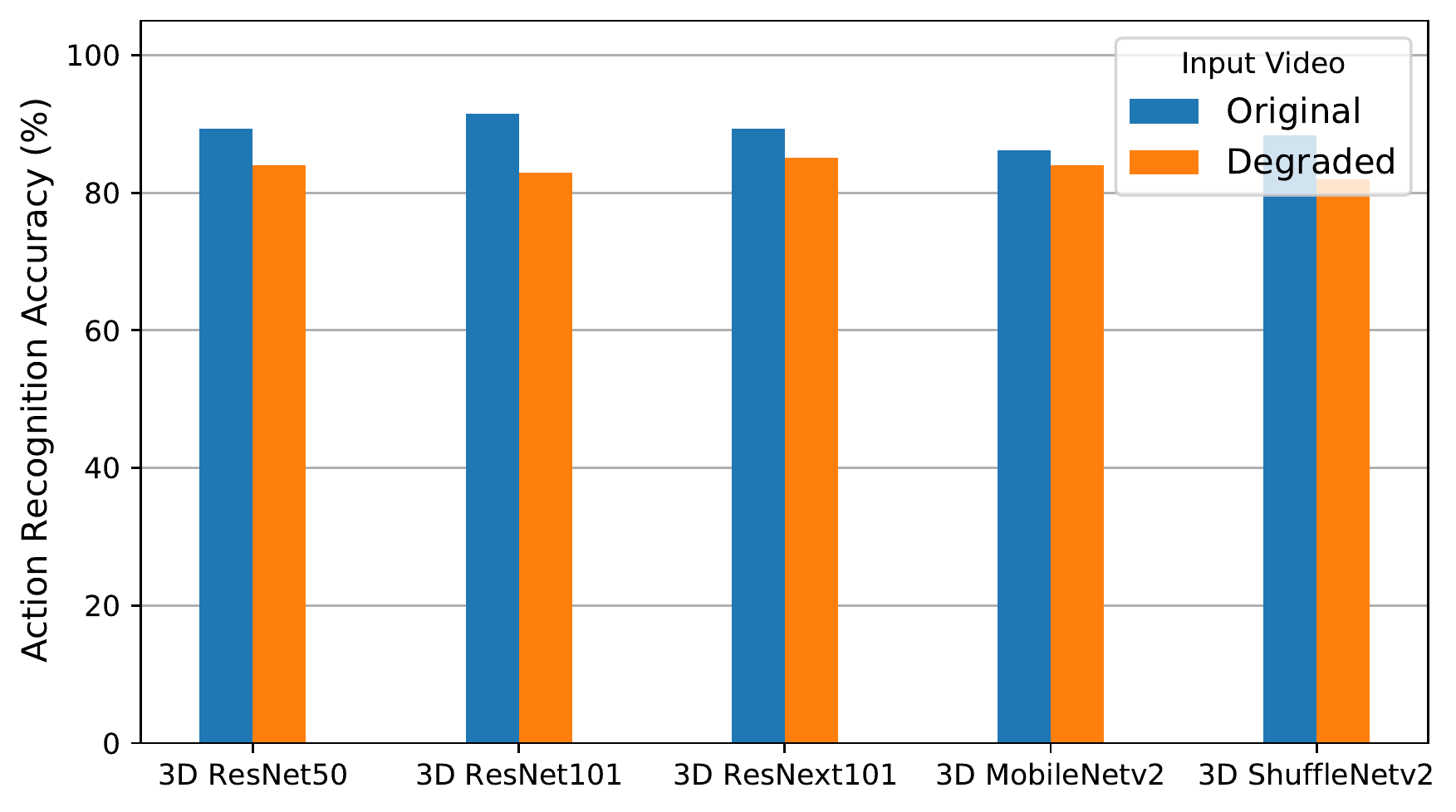}
\caption{Action recognition accuracy on various action recognition networks. \label{fig:analysis_utility} }
\end{figure}

In addition to strong privacy protection, a privacy-preserving model must allow task-specific features to be learned by any network that is designed for that task. In order to show that our proposed framework  allows useful spatio-temporal cues for action recognition,  we take the pre-trained BDQ encoder from Section~\ref{sec:experiments} and use its output to separately train five 3D CNNs for predicting action classes on the SBU dataset, as shown in Figure~\ref{fig:analysis_utility}. Furthermore, similar to Section~\ref{sec:analysis_privacy}, we also train these networks on the original videos to prepare corresponding baselines for comparison.  Note that all the networks are initialized with Kinetics-400 pre-trained weights. From Figure~\ref{fig:analysis_utility}, we observe that the BDQ encoder consistently allows spatio-temporal information to be learned with 3D ResNext-101 performing the best at 85.1\% and 3D ShuffleNet-v2 performing the worst at 81.91\%. Furthermore, all the networks achieve action recognition accuracy marginally lower than their corresponding baselines. 

\subsection{Robustness to Reconstruction Attack}\label{sec:analysis_reconst}
\begin{figure}[!tb]
     \centering
     \begin{subfigure}[b]{.15\textwidth}
         \centering
         \includegraphics[width=\textwidth]{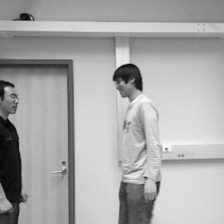}
         \includegraphics[width=\textwidth]{images/reconst/21_x_orig-0.png}
         \caption*{\scriptsize Original \\Frame }
     \end{subfigure}
     \hspace{-.5em}
     \begin{subfigure}[b]{.15\textwidth}
         \centering
         \includegraphics[width=\textwidth]{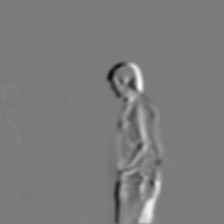}
         \includegraphics[width=\textwidth]{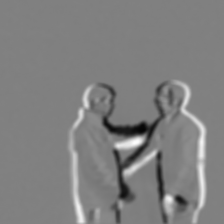}
         \caption*{\scriptsize BDQ \\($\alpha=0$)}
     \end{subfigure}
          \hspace{-.5em}
     \begin{subfigure}[b]{.15\textwidth}
         \centering
         \includegraphics[width=\textwidth]{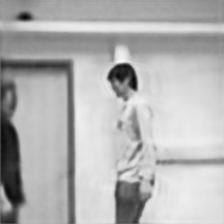}
         \includegraphics[width=\textwidth]{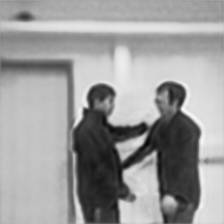}
         \caption*{\scriptsize Rec. BDQ \\($\alpha=0$)}
     \end{subfigure}
          \hspace{-.5em}
     \begin{subfigure}[b]{.15\textwidth}
         \centering
         \includegraphics[width=\textwidth]{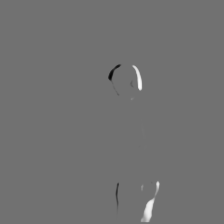}
         \includegraphics[width=\textwidth]{images/reconst/21_x_quant-0.png}
         \caption*{\scriptsize BDQ \\($\alpha=2$)}
     \end{subfigure}
          \hspace{-.5em}
     \begin{subfigure}[b]{.15\textwidth}
         \centering
         \includegraphics[width=\textwidth]{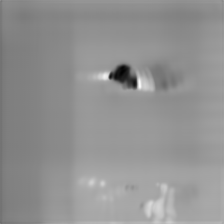}
         \includegraphics[width=\textwidth]{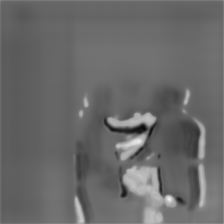}
         \caption*{\scriptsize Rec. BDQ \\($\alpha=2$)}
     \end{subfigure}
        \caption{Visualization of reconstruction results for  $\alpha=0$ and $2$. \label{fig:analysis_reconst}}
\end{figure}

In this section, we explore a scenario where an attacker has access to the BDQ encoder such that he/she can produce a large training set containing degraded videos along with their corresponding original videos. In such a case, the attacker can train an encoder-decoder network and try to reverse the effect of BDQ, recovering the privacy information. In order to show that our proposed framework is resistant to such an attack, we train a 3D UNet~\cite{cciccek20163d} model for 200 epochs on the SBU dataset with input as degraded videos from the pre-trained BDQ encoder of Section~\ref{sec:experiments} and output original videos as ground-truth labels. Additionally, we also train the 3D UNet model with input as degraded videos from an untrained BDQ encoder. Figure~\ref{fig:analysis_reconst} visualizes some examples of  reconstruction when videos from untrained (column 4) and trained (column 5) BDQ encoder are used for training. We observe that the reconstruction network can successfully reconstruct the original video with satisfactory accuracy when the input is from an untrained BDQ encoder. However, when the input is from a trained BDQ encoder, reconstruction is significantly poor and privacy information is still preserved.    

\subsection{Subjective Evaluation}
In Section~\ref{sec:analysis_privacy} and~\ref{sec:analysis_reconst}, we show that our proposed framework is robust against adversaries that may try to learn or reconstruct the privacy attributes. However, such protection is of no use unless it provides visual privacy against the human visual system. In order to show that the BDQ encoder produces encodings that provide visual privacy, we conduct a user study on the videos from the SBU dataset. The user study is composed of 60 questions where each question consists of a video sampled from the SBU validation set and applied with the BDQ encoder learned in Section~\ref{sec:experiments}. Furthermore, each question has seven options showing cropped faces of actors from the SBU dataset. Given a BDQ output video where two persons are interacting, the task of the user is to select the identities of both the actors from the seven options. A total of 26 participants took part in the study. Note that the random chance of selecting two actors and both of them being correct is 4.76\%. Similarly, the random chance of selecting two actors and at least one of them being correct is 52.38\%. 
In the first case, the results of our user study reveal that the participants were able to recognize both the actors correctly with an accuracy of 8.65\%. Similarly, the users' accuracy for the second case was 65.64\% (more details in supplementary).

\subsection{Comparison with Event Camera}
\begin{figure}[t]
     \centering
     \begin{subfigure}[b]{.14\textwidth}
         \centering
        \frame{\includegraphics[width=\textwidth]{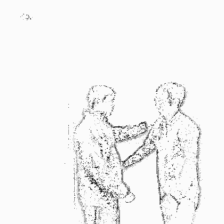}}
         \caption*{\scriptsize $th=0.4$ \\93.54\% \\73.99\%}
     \end{subfigure}
          \hspace{.1em}
     \begin{subfigure}[b]{.14\textwidth}
         \centering
         \frame{\includegraphics[width=\textwidth]{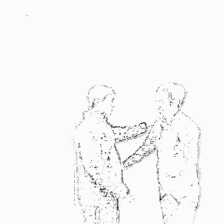}}
         \caption*{\scriptsize $th=0.8$ \\92.47\% \\58.33\%}
     \end{subfigure}
          \hspace{.1em}
     \begin{subfigure}[b]{.14\textwidth}
         \centering
         \frame{\includegraphics[width=\textwidth]{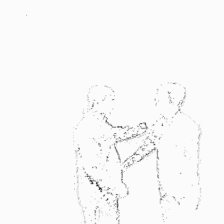}}
         \caption*{\scriptsize $th=1.2$ \\90.32\% \\47.84\%}
     \end{subfigure}
          \hspace{.1em}
     \begin{subfigure}[b]{.14\textwidth}
         \centering
         \frame{\includegraphics[width=\textwidth]{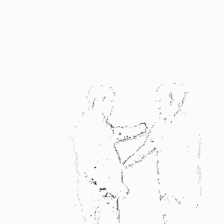}}
         \caption*{\scriptsize $th=1.6$ \\87.09\% \\46.23\%}
     \end{subfigure}
        \hspace{.1em}
     \begin{subfigure}[b]{.14\textwidth}
         \centering
        \frame{\includegraphics[width=\textwidth]{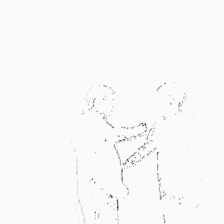}}
         \caption*{\scriptsize $th=2.0$ \\86.02\% \\40.12\%}
     \end{subfigure}
         \hspace{.1em}
     \begin{subfigure}[b]{.14\textwidth}
         \centering
         \frame{\includegraphics[width=\textwidth]{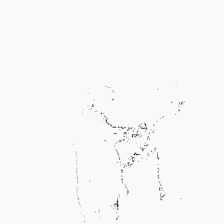}}
         \caption*{\scriptsize $th=2.4$ \\82.79\% \\34.87\%}
     \end{subfigure}
        \caption{Example event frames (Row 1), event threshold (Row 2), action recognition accuracy (Row 3) and actor-pair recognition accuracy (Row 4) on SBU. \label{fig:analysis_event}}
\end{figure}

In recent years, Dynamic Vision Sensor (DVS) based cameras are being proposed as an in-home monitoring solution for privacy-preserving action detection and recognition \cite{samsung}. Unlike traditional cameras that capture high-resolution videos and images, a DVS sensor detects the temporal changes in the pixel intensity at a pixel location. If the pixel intensity rises beyond a fixed threshold at a pixel, it is registered as a positive event. However, if it drops below a fixed threshold, it is registered as a negative event. Here, positive and negative events describe the direction of motion which can be encoded as outlines. At an abstract level, our BDQ encoder can be seen as a digital approximation of the DVS sensor that can be implemented in any traditional camera. In order to compare our framework with the DVS sensor, we first convert the videos from the SBU dataset into events using \cite{gehrig2020video} which proposes a method for converting any video into synthetic events such that they can be simulated as events from a DVS sensor. Such a method is useful since training DNNs require huge amount of data and collecting such amount is not feasible using event cameras. Furthermore, the above method enables us to vary the pixel level threshold which decides if the intensity change (event) will be registered or not. Note that a high threshold leads to less events while a low threshold leads to more events being registered. Figure~\ref{fig:analysis_event} displays the effect of threshold on an event frame.  The events from the converted SBU dataset are first converted into event frames using \cite{gehrig2020video} which are then used for training a 3D ResNet-50 model for action recognition and a 2D ResNet-50 model for actor-pair recognition. The initialization and training settings is identical to Section~\ref{sec:experiments}. Figure~\ref{fig:analysis_event} reports the trade-off for each threshold. We observe that with the increase in threshold, both action recognition and actor-pair accuracy drops. Furthermore, the trade-off at threshold value 2.4 is close to the trade-off achieved by the BDQ encoder with $\alpha=2$ (refer Section~\ref{sec:results}).

\section{Conclusions}
This paper proposes a novel encoder called BDQ for the task of privacy-preserving human action recognition. The BDQ encoder is composed of three modules: \textit{Blur}, \textit{Difference}, and \textit{Quantization} whose parameters are learned in an end-to-end fashion via an adversarial training framework such that it learns to allow important spatio-temporal attributes for action recognition and inhibit spatial privacy attributes. We show that the proposed encoder achieves state-of-the-art trade-off on three benchmark datasets in comparison to previous works. Furthermore, the trade-off achieved is at par with the DVS sensor-based event cameras. Finally, we also provide an extensive analysis of the BDQ encoder including an ablation study on its components, robustness to various adversaries, and a subjective evaluation.

\noindent\textbf{Limitation:} Due to its design, our proposed framework for privacy preservation cannot work in cases when the subject or the camera does not move.\\
\noindent\textbf{Acknowledgement:} This work was supported by JSPS KAKENHI Grant Number JP20K20628.

%
%

\end{document}